\documentclass[letterpaper]{article} 
\usepackage{aaai25}  
\usepackage{times}  
\usepackage{helvet}  
\usepackage{courier}  
\usepackage[hyphens]{url}  
\usepackage{graphicx} 
\usepackage{natbib}  
\usepackage{caption} 
\usepackage{algorithm}
\usepackage{algorithmic}

\usepackage{framed}
\usepackage{amssymb}
\usepackage{latexsym}
\usepackage{amsmath}
\usepackage{multirow}
\usepackage{booktabs}
\usepackage{tabularx}
\usepackage{caption}
\usepackage{colortbl}
\usepackage{xcolor}
\usepackage{csquotes}

\definecolor{lightgray}{gray}{0.9}

\usepackage{newfloat}
\usepackage{listings}
\DeclareCaptionStyle{ruled}{labelfont=normalfont,labelsep=colon,strut=off} 
\floatstyle{ruled}
\newfloat{listing}{tb}{lst}{}
\floatname{listing}{Listing}
\usepackage{cuted} 

\title{PointDGMamba: Domain Generalization of Point Cloud Classification  \\
via Generalized State Space Model}

\author{
    Hao Yang$^{1}$\footnotemark[1], 
    Qianyu Zhou$^{1}$\footnotemark[1], 
    Haijia Sun$^2$\thanks{\textit{These authors contributed equally.}},
    Xiangtai Li$^{3, 5}$, 
    Fengqi Liu$^{1}$,
    Xuequan Lu$^4$,\\
    Lizhuang Ma$^1$\thanks{\textit{Corresponding author.}},
    Shuicheng Yan$^{3, 5}$
}

\affiliations{
    \textsuperscript{\rm 1}Department of Computer Science and Engineering, Shanghai Jiao Tong University, Shanghai, China\\
    \textsuperscript{\rm 2}School of Information Management, Nanjing University, Nanjing, China\\
    \textsuperscript{\rm 3}Skywork AI, Singapore\\
    \textsuperscript{\rm 4}Department of Computer Science and Software Engineering, The University of Western Australia, Australia\\
    \textsuperscript{\rm 5}Nanyang Technological University, Singapore\\
    
    \{yanghao.cs, zhouqianyu, liufengqi, lzma\}@sjtu.edu.cn, 522023140081@smail.nju.edu.cn\\

}

\usepackage{bibentry}

\begin{document}

\maketitle

\begin{figure*}[t]
    \centering

    \includegraphics[scale=0.466]{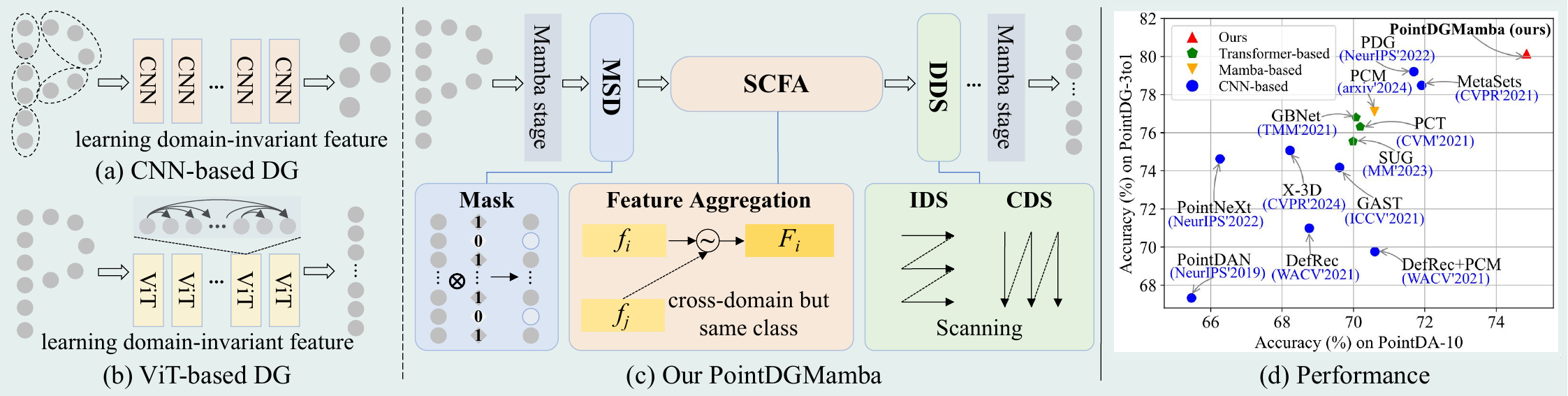}

    \captionof{figure}{\textbf{Left:} Comparisons between previous works and our PointDGMamba. Previous domain generalization (DG)-based point cloud classification (PCC) methods often rely on Convolution Neural Networks (CNNs) or Vision Transformers (ViTs) to learn domain-invariant features, while often suffering from limited receptive fields with local perception (a) or high quadratic complexity with global perception (b). \textbf{Middle:} In contrast, we propose a novel framework, PointDGMamba (c), that excels in strong generalizability toward unseen domains and has the advantages of global receptive fields and efficient linear complexity. Our PointDGMamba consists of Masked Sequence Denoising (MSD), Sequence-wise Cross-domain Feature Aggregation (SCFA), and Dual-level Domain Scanning (DDS).
    \textbf{Right:} In the widely-used PointDA-10 benchmark and our proposed PointDG-3to1 benchmark (d), our PointDGMamba demonstrates superior accuracy against state-of-the-art methods. 
    }
    \label{teaser}

\end{figure*}

\begin{abstract}

Domain Generalization (DG) has been recently explored to improve the generalizability of point cloud classification (PCC) models toward unseen domains. However, they often suffer from limited receptive fields or quadratic complexity due to using convolution neural networks or vision Transformers. In this paper, we present the first work that studies the generalizability of state space models (SSMs) in DG PCC and find that directly applying SSMs into DG PCC will encounter several challenges: the inherent topology of the point cloud tends to be disrupted and leads to noise accumulation during the serialization stage. Besides, the lack of designs in domain-agnostic feature learning and data scanning will introduce unanticipated domain-specific information into the 3D sequence data. To this end, we propose a novel framework, \textit{PointDGMamba}, that excels in strong generalizability toward unseen domains and has the advantages of global receptive fields and efficient linear complexity. PointDGMamba consists of three innovative components: Masked Sequence Denoising (MSD), Sequence-wise Cross-domain Feature Aggregation (SCFA), and Dual-level Domain Scanning (DDS). In particular, MSD selectively masks out the noised point tokens of the point cloud sequences, SCFA introduces cross-domain but same-class point cloud features to encourage the model to learn how to extract more generalized features. DDS includes intra-domain scanning and cross-domain scanning to facilitate information exchange between features. In addition, we propose a new and more challenging benchmark PointDG-3to1 for multi-domain generalization. Extensive experiments demonstrate the effectiveness and state-of-the-art performance of PointDGMamba. 
\begin{links}
\link{Code}{https://github.com/yxltya/PointDGMamba}
\end{links}

\end{abstract}

\section{Introduction}

3D point cloud learning is a fundamental 3D vision task that has applications in many fields, such as autonomous driving, robot navigation, virtual reality, and urban modeling. Some methods such as PointNet~\cite{qi2017pointnet}, PointNet++~\cite{qi2017pointnet++}, and DGCNN~\cite{phan2018dgcnn} have made great progress in point cloud classification. 
However, existing methods typically excel only on seen datasets and may encounter performance degradation on unseen domains. This is mainly due to domain shifts, \emph{e.g.,} differences caused by sensor types, scanning angles, and environmental conditions, across different domains.

To address this issue, domain generalization (DG) techniques have been recently introduced into point cloud classification (PCC)~\cite{huang2021metasets} to learn domain-invariant features, thereby improving the model's generalizability.
The mainstream methods include data augmentation~\cite{xiao20233d}, adversarial training~\cite{lehner20223d}, and consistency learning~\cite{kim2023single}. Nevertheless, most of them are based on CNNs and inherently suffer from a limited receptive field, making it difficult to capture global information. As a result, the model may struggle to fully understand the overall structure of the data, leading to suboptimal classification performance. One solution is to use a Vision Transformer~\cite{han2022survey}, but its internal attention layer inevitably introduces higher quadratic computational complexity with increased input points. 
Therefore, how to effectively model global information while keeping low computational complexity is a key issue in DG PCC.

Recently, Mamba~\cite{gu2023mamba}, as an emerging State Space Model (SSM) model, has been demonstrated to be effective in capturing long-range dependencies and global information~\cite{xing2024segmamba,zhang2024motion}. More importantly, they can accomplish some tasks with linear complexity, which solves the problem of a limited receptive field and avoids quadratic complexity. The representative work, Point Mamba~\cite{liang2024pointmamba} and Point Cloud Mamba (PCM)~\cite{zhang2024point} propose several serialization methods to scan 3D point cloud data into specific sequences. Nonetheless, these methods often produce less desirable performance in unseen domains because they do not adequately account for domain shifts or do not include any tailored designs. 
As shown in Figure~\ref{teaser}, there exist noticeable performance gaps between PCM and state-of-the-art DG PCC methods. Thus, 
 and understanding the barriers preventing Mamba-based models from effectively handling distribution shifts in DG and enhancing the generalizability of SSMs
 remain critical challenges for the point cloud field.

In this paper, we aim to boost the generalizability of Mamba-like models toward unseen domains in point cloud classification. Our motivations mainly lie in \textit{three aspects}. \textit{Firstly}, we observe that the inherent topology of the point cloud tends to be disrupted during Mamba's serialization process, and even generates some unexpected noises unrelated to the current state. Such noise would accumulate during the training, subsequently affecting the model's performance when unseen data is used as input. \textit{Secondly}, we observe that existing blocks of Mamba-like models are usually hand-crafted and over-heuristic, overlook the designs of learning domain-agnostic features, and tend to overfit specific domains. This may introduce unanticipated domain-specific information into the sequence data, thereby weakening Mamba's effectiveness in handling distribution shifts. \textit{Thirdly}, an unresolved issue is how to effectively convert 3D point cloud data into 1D sequence data suitable for Mamba in DG PCC. Though recent studies have investigated different scanning methods for point cloud tasks, these rigid and fixed scanning approaches inevitably introduce human biases and largely ignore domain-agnostic considerations. Additionally, they are highly susceptible to varying conditions, posing challenges to their application in unseen domains.

Motivated by the above facts, we propose PointDGMamba, a novel State Space Model-based framework for domain generalizable point cloud classification.
PointDGMamba excels in strong generalizability toward unseen domains and has the advantages of global receptive fields and efficient linear complexity. 
It mainly includes three core components: Masked Sequence Denoising (MSD), Sequence-wise Cross-domain Feature Aggregation (SCFA), and Dual-level Domain Scanning (DDS).
Specifically, MSD selectively masks the noised point tokens of the point cloud sequences and uses the purified features for classification, mitigating the adverse effects of noise accumulation.
Then, SCFA is designed to aggregate cross-domain but same-class point cloud features to prompt the model to extract more domain-generalized features. 
Finally, DDS, including two scanning methods, intra-domain scanning and cross-domain scanning, is proposed to facilitate sufficient information interaction between different parts of the features. 
As such, it converts 3D point cloud data into 1D sequence data suitable for Mamba-like models in varying unseen domains.
Provided that the number of source domains in the existing DG PCC benchmark is limited, we also present a new multi-domain generalization benchmark, PointDG-3to1, which is more diverse, practical, and challenging. It includes 4 variants of leave-one-out settings, with 3 domains used as source domains and the remaining one as the unseen domain.

Our contributions can be summarized as follows:

\begin{itemize}
    \item We propose PointDGMamba, a novel State Space Model-based framework for domain generalizable point cloud classification that shows strong generalizability toward unseen domains and has the advantages of global receptive fields and efficient linear complexity.

    \item We design Masked Sequence Denoising (MSD), Cross-domain Feature Aggregation (SCFA), and Dual-level Domain Scanning(DDS) to improve the generalizability of the SSM-based model in DG PCC.

    \item We propose a more challenging multi-domain generalization benchmark PointDG-3to1. Extensive experiments on PointDA-10 and our PointDG-3to1 benchmarks demonstrate the effectiveness and superiority of our PointDGMmaba against state-of-the-art competitors.

\end{itemize}

\begin{figure*}[!ht]
\centering

\includegraphics[scale=0.50]{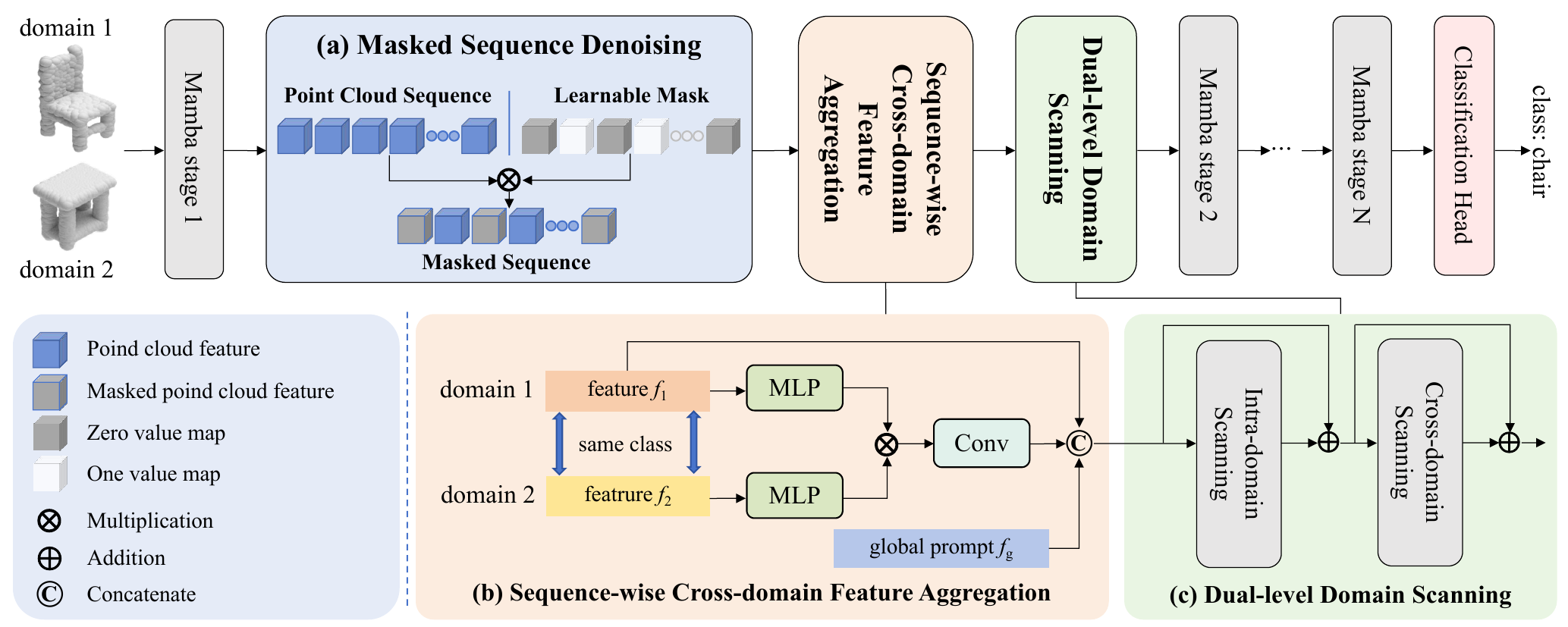}

\caption{The framework of PointDGMamba. It consists of three components: (a) Masked Sequence Denoising (MSD) is presented to mask out noised point patches in the sequence and thus mitigate adverse effects of noise accumulation during the serialization stage; (b) Sequence-wise Cross-domain Feature Aggregation (SCFA) is introduced to aggregate cross-domain but same-class point cloud features with the global prompt to extract more generalized features, thereby strengthening Mamba's effectiveness in handling distribution shifts. (c) Dual-level Domain Scanning, including intra-domain scanning and cross-domain scanning, is proposed to facilitate sufficient information interaction between different parts of the features.}

\label{1-network}
\end{figure*}

\section{Related Work}

\noindent {\bf{Point Cloud Classification}} (PCC) makes an accurate classification of the given 3D point cloud data. 
Pioneering works~\cite{qi2017pointnet,qi2017pointnet++} use MLP-like architecture to learn the representation on the point cloud directly. Researchers have explored many frameworks~\cite{li2018pointcnn,wang2019dynamic} based on Convolutional Neural Networks (CNNs) to enhance the understanding of point cloud geometric structures.
However, they suffer from a limited receptive field when stacking deep layers. Recently, vision Transformers~\cite{zhao2021point,fang2024explore,deng2024vg4d} have been introduced to PCC due to the merit of global receptive fields inherent in ViT. Some representative methods, such as PCT~\cite{guo2021pct} and Point Transformer~\cite{zhao2021point}, unveil the potential of self-attention layers and
propose Transformer-based architectures to model global context and dependencies within point clouds. Point-BERT \cite{yu2022point}, and Point-MAE \cite{pang2022masked} introduce the Masked Point Modeling (MPM) as a pre-text task for reconstructing masked point clouds. They use masked autoencoders for BERT-style pre-training or self-supervised learning to achieve stronger representations. Despite their gratifying progress, most of these methods perform well on seen datasets and may encounter significant performance degradation when generalizing to unseen, novel domains.

\noindent {\bf{Domain Generalized Point Cloud Classification.}}
Although domain adaptation techniques have been explored in point cloud areas~\cite{liu2024cloudmix,jiang2024pcotta,li2024dapointr}  to narrow the domain shifts, the target data is not always accessible in real scenarios, which might fail these methods.
Domain generalization has recently been introduced into PCC to improve the generalizability toward unseen domains. The mainstream DG PCC models aim to learn domain-invariant features and are mainly categorized as adversarial learning, consistency learning~\cite{kim2023single}, data augmentation~\cite{lehner20223d,xiao20233d}, \emph{etc}. Despite their remarkable progress in DG, the lack of global receptive fields hinders further developments of CNN-based models for boosting generalization performance. 
To address this issue, \textit{Huang et al.}~\cite{huang2023sug} propose subdomain alignment and domain-aware attention to achieve DG with Transformers. Nonetheless, it suffers from quadratic complexity to the resolution rising from the attention mechanism, leading to extra computation and memory overhead. Moreover, it is limited to single-source domain generalization. Thus, it is necessary to investigate the multi-domain generalization that excels in global information modeling and low computational complexity in PCC.

\noindent {\bf{Mamba.}} Mamba~\cite{gu2023mamba}, as well as the state space model (SSM), has garnered increasing attention due to its significant advantages in global receptive fields and computational complexity. VMamba~\cite{liu2024VMamba} and Vim~\cite{zhu2024vision} propose visual SSMs to deploy Mamba for vision tasks. In the area of point cloud understanding, PointMamba~\cite{liang2024pointmamba} introduced a reordering strategy to scan data in a specific sequence to capture point cloud structures. Besides, Mamba3D~\cite{han2024mamba3d} used local norm pooling blocks to extract local geometric features. 
Similarly, \textit{Zhang et al}. presented PCM~\cite{zhang2024point}, which converts point clouds into one-dimensional point sequences using a consistent traversal serialization method, ensuring that adjacent points in the sequence remain adjacent in space. 
Unfortunately, no research investigates Mamba's generalizability in point cloud tasks. 
To our knowledge, this is the first work that studies the generalizability of SSM-based models toward unseen domains in point cloud tasks. This paper uses the popular PCM~\cite{zhang2024point} as the baseline. 

\section{Method}

In this section, we present PointDGMamba, a novel State Space Model-based framework for DG point cloud classification that
excels in strong generalizability toward unseen domains and has the advantages of global receptive fields and efficient linear complexity. As shown in Figure~\ref{1-network}, it includes three core components: Masked Sequence Denoising (MSD), Sequence-wise Cross-domain Feature Aggregation (SCFA), and Dual-level Domain Scanning(DDS). Concretely, MSD is presented to selectively mask out noised point patches in the sequence and thus alleviate the noise accumulation during the serialization stage. Besides, SCFA is introduced to aggregate cross-domain but same-class point cloud features with the global prompt to extract more generalized features, thereby strengthing Mamba's effectiveness in handling distribution shifts. Finally, DDS, including two scanning methods: intra-domain scanning and cross-domain scanning, is proposed to facilitate sufficient information interaction between different parts of the features. All modules are inserted after the first Mamba stage.

\subsection{Masked Sequence Denoising}

We have noticed that during Mamba's serialization process, the inherent topology of the point cloud often gets disrupted when 3D point cloud data is converted into 1D sequences, leading to the generation of unexpected noise that is unrelated to the current state. This noise can be accumulated during the training, which could negatively impact the model’s performance when it encounters unseen data.

To address this issue, we propose Masked Sequence Denoising (MSD) to selectively mask the noised point tokens of the point cloud sequences and use the purified features for classification, mitigating the adverse effects of noise accumulation. This process not only preserves the basic feature of the point cloud but also ensures that the denoised sequence can highly represent the original structure. Specifically, we define the point cloud feature as $f$ and the mask as $m$. The masked feature map $F$ can be represented as:
\begin{equation}
\begin{aligned}
\label{eq-a-1}
F & = f \otimes m,
\end{aligned}
\end{equation}
Ideally, an element of 0 in the mask sequence indicates that the features are to be masked, while an element of 1 indicates that the features are to be preserved. However, since this mask consists of only 0 and 1 values, the gradient cannot be back-propagated during the training. We can solve this problem by using the Gumbel-Softmax method~\cite{jang2016categorical}, where the mask $m$ at a certain position can be represented as:
\begin{equation}
\begin{aligned}
\label{eq-a-2}
m &= \frac{\exp   \left(\frac{  g_{1} + \log p_{1} }{\tau}\right)   }{\sum\limits_{i=0}^{1} \exp\left(\frac{ g_{i} + \log p_{i} }{\tau}\right)}
\end{aligned}
\end{equation}

\noindent where \( g_{i} \) represents Gumbel noise, \(\tau\) is the temperature parameter, and \( p_{i} \) represents the probability that the current position is 0 or 1. The higher the probability of \(p_{i}\) becomes, the closer the probability value of the current position is to a value of 1, otherwise it is closer to a value of 0. This indicates that while features are preserved, noise is greatly suppressed. As shown in Figure~\ref{1-network}(a), the input sequence will be multiplied by a learnable mask to obtain the masked sequence. Then, the denoised and purified sequence features will be forwarded for classification.

\noindent \textbf{Remark.} Unlike existing Masked Point Modeling (MPM)-based methods, we intend to filter out the noised tokens selectively rather than reconstruct the masked tokens in the point cloud sequence. Besides, as for the technical designs, our MSD involves addressing the challenge of learning a discrete number (0 or 1) during the back-propagation, setting it apart from other MPM frameworks.

\subsection{Sequence-wise Cross-domain Feature Aggregation}
\label{sec1}

We observe that the current blocks of Mamba models are manually designed and over-heuristic. They neglect the designs of domain-agnostic features and tend to overfit specific domains. This can introduce unexpected domain-specific information into the sequence data, reducing Mamba's effectiveness in dealing with distribution shifts.

To address this, we propose a Sequence-wise Cross-domain Feature Aggregation (SCFA) to aggregate cross-domain but same-class point cloud features to prompt the model to extract more generalized features.  
Specifically, as shown in Figure~\ref{1-network}(b), the denoised point cloud features $f_1$ are fed into this module. Then, the randomly selected same-class point cloud feature from other domains $f_2$ is aggregated with $f_1$ to get the Cross-Domain Feature $f^{’}$:
\begin{equation}
\begin{aligned}
\label{eq-C-1}
f^{’} & = \mathrm{Conv} (\mathrm{MLP} (f_1) \otimes \mathrm{MLP} (f_2)),
\end{aligned}
\end{equation}

\noindent where $\otimes$ represents the element-wise multiplication, $\mathrm{MLP}$ represents multi-layer perceptron and $\mathrm{Conv}$ represents a convolutional layer.

In addition, we also introduce Global Prompt $f_g$ to capture global information of the entire source domain, which consists of a set of learnable vectors. It is similar to a system message in large language models. Then, we aggregate the global prompt $f_g$ with $f_1$ and $f^{’}$ to avoid unanticipated domain-specific information in the sequence data. During the training, point cloud features will be re-aggregated as:
\begin{equation}
\begin{aligned}
\label{eq-C-1}
F & = \mathrm{Concat} (f_1,f^{’},f_g), 
\end{aligned}
\end{equation}

\noindent where $F$ is the aggregated point cloud features.

\begin{figure}[t]
\centering
\includegraphics[scale=0.70]{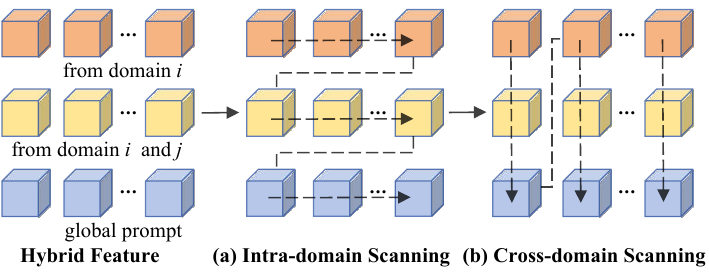}

\caption{Dual-level Domain Scanning (DDS) comprises Intra-domain Scanning and Cross-domain Scanning.}

\label{2-network}
\end{figure}

\subsection{Dual-level Domain Scanning}
\label{sec2}

A key challenge is converting 3D point cloud data into 1D sequence data suitable for Mamba in DG PCC. While recent studies have explored various scanning methods for point cloud tasks, these rigid and fixed approaches often introduce human biases and largely overlook domain-agnostic factors. Moreover, they are susceptible to varying conditions, making it difficult to apply them in unseen domains.

In order to facilitate the interaction of different feature information for generalization, we design Dual-level Domain Scanning, including Intra-domain Scanning (IDS) and Cross-domain Scanning (CDS). As shown in Figure~\ref{2-network}. The cubes of different colors in the figure represent different features. IDS treats features as three unrelated sequences, scanning them one after another in order. Only after scanning the current feature can the next feature be scanned. CDS treats features as three related sequences. After scanning a data point of the first feature, the data points at the same position for the other two features will be scanned sequentially. This can promote the interaction between each feature. We use arrows in the figure to indicate the order in which each data is scanned. The scanning process could be formulated as:

\begin{equation}
\begin{aligned}
\label{eq-D-1}
F_{out} & = \mathrm{CDS} (\mathrm{IDS} (F)),
\end{aligned}
\end{equation}

\noindent where $F_{out}$ is the output feature after two types of scanning. Then, the output features will be sent to the later Mamba stages and classification head.
It is worth noting that the DDS module processes aggregated features, so it needs to be used together with the SCFA module.

\begin{table*}[t]
    \renewcommand\arraystretch{1.15} 
		\setlength{\tabcolsep}{0.40mm}

		\begin{center}
  \resizebox{1.0\textwidth}{!}{%
		  \begin{tabular}
{c|c| c|c| c c c >{\columncolor{lightgray}}c |cccc>{\columncolor{lightgray}}c}

\hline

\multirow{2}{*}{\bf{Method}} & \multirow{2}{*}{\bf{Setting}} & \multirow{2}{*}{\bf{Venue}} & \multirow{2}{*}{\bf{Backbone}} & \multicolumn{4}{c|}{ \textit{PointDA-10 Benchmark}} & \multicolumn{5}{c}{\textit{PointDG-3to1 Benchmark}} \\

 & & & & \small{M,S*→S} & \small{M,S→S*} & \small{S,S*→M} & \bf{Avg.} & \small{ABC→D} & \small{ABD→C} & \small{ACD→B} & \small{BCD→A} & \bf{Avg.} \\

\hline
PointDAN & DA & NeurIPS'2019 & PointNet & 77.38 & 40.32 & 78.69 & 65.46 & 58.85 & 81.66 & 48.86 & 79.95 & 67.33  \\
DefRec & DA & WACV'2021 & DGCNN & 77.23 & 44.28 & 84.77 & 68.76 & 72.76 & 79.97 & 43.29 & 87.94 & 70.99  \\
DefRec+PCM & DA  & WACV'2021 & DGCNN & 80.02 & 48.39 & 83.39 & 70.60 & 68.81 & 82.90 & 44.00 & 83.33 & 69.76  \\
GAST & DA & ICCV'2021 &  DGCNN & 79.43 & 47.69 & 81.72 & 69.61 & 71.78 & 86.43 & 52.31 & 86.21 & 74.18  \\
\hline
MetaSets & DG  & CVPR'2021 & PointNet & 81.39 & 50.86 & 83.48 & 71.91 & 73.24 & 92.41 & 60.97 & 87.28 & 78.48  \\
PDG & DG & NeurIPS'2022 &  PointNet & 79.82 & 51.73 & 83.51 & 71.69 & 73.38 & 92.98 & 60.57 & 89.90 & 79.21  \\
PointNeXt & DG & NeurIPS'2022 & PointNet & 77.31 & 43.32 & 78.16 & 66.26 & 71.47 & 91.70 & 46.39 & 88.95 & 74.63 \\

X-3D & DG & CVPR'2024 & PointNet & 78.06 & 46.91 & 79.69 & 68.22 & 71.58 & 91.89 & 48.34 & 88.45 & 75.07 \\

PCT & DG  & CVM'2021 & PointTrans & 80.23 & 48.29 & 81.91 & 70.14 & 71.43 & 87.43 & 58.43 & 88.34 & 76.41  \\
GBNet & DG & TMM'2021 & PointTrans & 79.94 & 48.92 & 81.34 & 70.07 & 72.78 & 87.83 & 57.76 & 88.82 & 76.80  \\
SUG & DG & MM'2023 &  PointTrans & 78.34 & 49.59 & 82.03 & 69.99 & 71.58 & 89.62 & 54.66 & 86.35 & 75.55  \\

PCM & DG  & AAAI'2025 & Mamba & 81.02 & 46.83 & 83.92 & 70.59 & 72.27 & 91.24 & 57.28 & 87.54 & 77.08 \\

\hline
PointDGMamba & DG  & - & Mamba & \textbf{84.33} & \textbf{52.83} & \textbf{87.38} & \textbf{74.85} & \textbf{74.20} & \textbf{95.51} & \textbf{61.71} & \textbf{90.68} & \textbf{80.53} \\
\hline

            \end{tabular}
            }
        \end{center}
        
\caption{Comparison results with state-of-the-art PCC methods on the PointDA-10 and our proposed PointDG-3to1 benchmark. \textbf{Avg.} denotes the average classification accuracy across all target domains. We have highlighted the best result in \textbf{black}.}

\label{tab-first}
\end{table*}

\subsection{Training and Inference}

\noindent {\bf{Training.}} During the training, we use Cross Entropy loss to measure the difference between the model's prediction and the ground truths. Specifically, for given input samples $x$ and its labels $y$, Cross Entropy loss $\mathcal{L}_{class}$ can be formulated as:
\begin{equation}
\begin{aligned}
\label{eq-E-1}
\mathcal{L}_{class} & = - \sum_{i} y_i \log (\hat{y}_i),
\end{aligned}
\end{equation}
\noindent where $\hat{y}_i$ is the probability that the sample $x_i$ predicted by the model belongs to a certain class.

In addition, our PointDGMamba model requires a set of point clouds with the same class but from different source domains during the training phase for each experiment. 
We employ random resampling techniques during the data loading to ensure that the number of same-class point clouds in different source domains is consistent.

\noindent {\bf{Inference.}} During the inference phase, only the point cloud data from a single domain is input into the model. It is worth noting that each point cloud does not need to interact with features from other domains except the Global Prompt. We replace it with each point cloud's feature itself in SCFA.

\section{Experiments}
\subsection{Experiment Settings}
\label{experiment1}
{\bf{Implementation.}} In the training process, we used the AdamW~\cite{loshchilov2017decoupled} optimizer with an initial learning rate of $1e^{-4}$, a cosine decay schedule, and a weight decay of $1e^{-4}$. The number of epochs was set to 200. During the first 5 epochs, we employed a warmup mechanism to gradually increase the learning rate, reducing initial instability. Subsequently, the learning rate decreased following a cosine function, maintaining a relatively low learning rate of $1e^{-5}$ in the later stages of training. During the training phase, we used the PointMix~\cite{chen2020pointmixup} method for data augmentation to obtain more training samples.

\noindent {\bf{Benchmark.}} To evaluate our method, we use the widely-used PointDA-10~\cite{qin2019pointdan} benchmark, which consists of ModelNet-10(M), ShapeNet-10(S), and ScanNet-10(S*), and contain 10 shared categories. Among them, ModelNet and ShapeNet were obtained from synthetic 3D models. ScanNet is sampled from the real world. Compared to ModelNet and ShapeNet, point clouds in ScanNet often have certain missing parts due to object occlusion. In this benchmark, we randomly select two domains as the source domain and the remaining domain as the target domain for each experiment, including 3 DG scenarios: (1) M, S*→S; (2) M, S→S* and (3)S, S*→M. In all experiments, both the training and testing sets of the source domains are used, while the target domain only uses the testing set.

\subsection{PointDG-3to1 Benchmark}

The existing DG PCC benchmarks usually include a limited number of source domains, \emph{e.g.,} the number of source domains is only two, resulting in a lack of sufficient object diversities in the training data.
This is not conducive to the model's generalization in the unseen domain. In this paper, we propose a multi-domain generalization benchmark named PointDG-3to1 consisting of four sub-datasets, which is more diverse, practical, and challenging.

The PointDG-3to1 benchmark includes four sub-datasets: ModelNet-5 (A), ScanNet-5(B), ShapeNet-5 (C), and 3D-FUTURE-Completion (D). There are 5 shared classes in each dataset, including \enquote{cabinet}, \enquote{chair}, \enquote{lamp}, \enquote{sofa}, and \enquote{table}. The 3D-FUTURE-Completion~\cite{liu2024cloudmix} was generated from the 3D-FUTURE~\cite{fu20213d} dataset, with each point cloud consisting of 2048 points. When performing DG, it includes 4 variants of leave-one-out (LOO) settings: (1) ABC→D; (2) ABD→C; (3) ACD→B, and (4) BCD→A, where 3 domains are used as source domains and the remaining one as the unseen domain. Following the common practice of DG, whole source domains are used for training (the training and testing sets), and the testing set is used for evaluation.

\subsection{Comparisons to the State-of-the-art Methods}

\noindent {\bf{Comparison Methods.}} To evaluate the effectiveness of our proposed method, we compare several state-of-the-art approaches in PCC. These include CNN-based methods such as PointDAN~\cite{qin2019pointdan}, DefRec~\cite{achituve2021self}, GAST~\cite{zou2021geometry}, PDG~\cite{wei2022learning}, MetaSets~\cite{huang2021metasets}, PointNeXt~\cite{qian2022pointnext} and X-3D~\cite{sun2024x}, and Transformer-based methods such as SUG~\cite{huang2023sug}, PCT~\cite{guo2021pct} and GBNet~\cite{qiu2021geometric}, and Mamba-based methods such as PCM~\cite{zhang2024point}. Since some Transformer-based and Mamba-based methods were not originally designed for domain generalization, we made appropriate modifications to enable their application to DG PCC settings. \textit{More details could be referred to the supplementary.}

\begin{table}[t]

    \renewcommand\arraystretch{1.02} 
		\setlength{\tabcolsep}{1.1mm}
		\begin{center}
			\begin{tabular}
{c| c|c| ccc|  >{\columncolor{lightgray}} c}

\hline
\multicolumn{2}{c|}{\bf{SCFA}} & \multirow{2}{*}{\bf{MSD}} & \multirow{2}{*}{M,S*→S} & \multirow{2}{*}{M,S→S*} & \multirow{2}{*}{S,S*→M} & \\
\cline{1-2}
\bf{CDF} & \bf{GP} &  &   &  &  & \multirow{-2}{*}{\bf{Avg.}}  \\
				
\hline
& & & 82.45 & 48.19 & 81.49 & 70.71 \\
\checkmark &  & & 83.35 & 50.07 & 84.83 & 72.75 \\
\checkmark & \checkmark &  & 83.51 & 50.72 & 85.46 & 73.23 \\
    
\hline
\checkmark & \checkmark & \checkmark & 84.33 & 52.83 & 87.38 & 74.85\\

\hline
\end{tabular}
\end{center}

\caption{Ablation studies on SCFA and MSD modules on the PointDA-10 benchmark. CDF and GP represent Cross-Domain Features and Global Prompt, respectively.}

    \label{tab2}
\end{table}

\noindent {\bf{Results on PointDA-10 benchmark.}} As shown in Table~\ref{tab-first}, we report the comparison results on the PointDA-10 benchmark, indicating that our PointDGMamba achieves the best generalization performance on all these domains. Specifically, Our PointDGMamba outperforms the existing state-of-the-art methods, \emph{e.g.,} PointDGMamba is superior to MetaSets by 2.94\% in average generalization performance, demonstrating the effectiveness of our PointDGMamba in boosting the generalizability. The main reason is that these DG methods heavily rely on CNNs and ViTs, either suffering from limited receptive fields or quadratic complexity. In contrast, our SSM-based method, PointDGMamba, achieves global receptive fields and linear complexity and makes tailored designs for learning domain-invariant features in SSM.

\noindent {\bf{Results on PointDG-3to1 benchmark.}}
As shown on the right of Table~\ref{tab-first}, we report the generalization performance on the PointDG-3to1 benchmark, and our proposed PointDGMamba achieves superior generalization performance in all scenarios. In particular, the average generalization performance of PointDGMamba exceeds the existing state-of-the-art method PDG by 1.32\%. On the challenging 3D Feature and ScanNet target domains, our method outperforms the second-best method by 0.82\% and 0.74\%, respectively, which also demonstrates the superiority of our PointDGMamba to CNN-based and ViT-based methods.

\begin{table}[t]

    \renewcommand\arraystretch{1.02} 
		\setlength{\tabcolsep}{1.9mm}
		\begin{center}
			\begin{tabular}
				{ c|c |ccc| >{\columncolor{lightgray}}c}
				\hline

                \multicolumn{2}{c|}{\bf{DDS}} & \multirow{2}{*}{M,S*→S} & \multirow{2}{*}{M,S→S*} & \multirow{2}{*}{S,S*→M} & \\
                \cline{1-2}
				\bf{IDS} & \bf{CDS} &  &  &  & \multirow{-2}{*}{\bf{Avg.}} \\
				
                \hline
				 & & 82.99 & 50.92 & 84.65 & 72.85 \\
                \checkmark & & 83.70 & 51.89 & 85.82 & 73.80 \\
                & \checkmark & 84.15 & 51.72 & 85.60 & 73.82 \\
    
				\hline
                \checkmark & \checkmark & 84.33 & 52.83 & 87.38 & 74.85\\
				\hline
			\end{tabular}
	\end{center}

\caption{Ablation studies on the DDS module on the PointDA-10 benchmark, where IDS and CDS are Intra-domain Scanning and Cross-domain Scanning, respectively.}

    \label{tab_ablation2}
\end{table}

\subsection{Ablation Study}
In this section, we provide ablation studies to verify the contribution of each proposed component. For simplicity, all experiments are conducted using the PointDA-10 benchmark.

 \noindent {\bf{Effectiveness of SCFA and MSD.}} 
 Table \ref{tab2} demonstrates the contribution of SCFA and MSD while keeping DDS unchanged.
As aforementioned, SCFA leverages three parts of point cloud features: source domain features, cross-domain features (CDF), and global prompt (GP). The first row means just using the source domain features without any aggregation. After adding CDF, the generalization performance is improved in all scenarios, proving the effectiveness of CDF in promoting the extraction of more generalized features in the model. After using GP, the performance has also been improved, indicating that the model can extract generalization features on a global scale. Finally, after further adding MSD, we achieved the best generalization performance, indicating that MSD has significant advantages in removing noise that is not conducive to generalization.

\noindent {\bf{Effect of IDS and CDS.}} Table~\ref{tab_ablation2} shows the impact of Intra-Domain Scanning (IDS) and Cross-Domain Scanning (CDS) of our DDS module. The baseline means 
the model without any scanning, and the accuracy is less desired, indicating that scanning operations are crucial for feature processing and information fusion. Specifically, when using IDS only, the model can achieve certain performance improvements because it can capture the relationships between features to some extent. However, due to the lack of cross-domain information interaction, the classification accuracy is still not perfect. When using only CDS, we can also observe improvements in performance since the model can better fuse feature information from different domains, while the performance is not the best 
compared to using both scanning methods simultaneously. 
This indicates that during the feature scanning process, we should focus on not only the feature information within the domain but also cross-domain feature interaction, to enhance the model's generalizability.

\begin{figure}[t]
\centering

\includegraphics[scale=0.45]{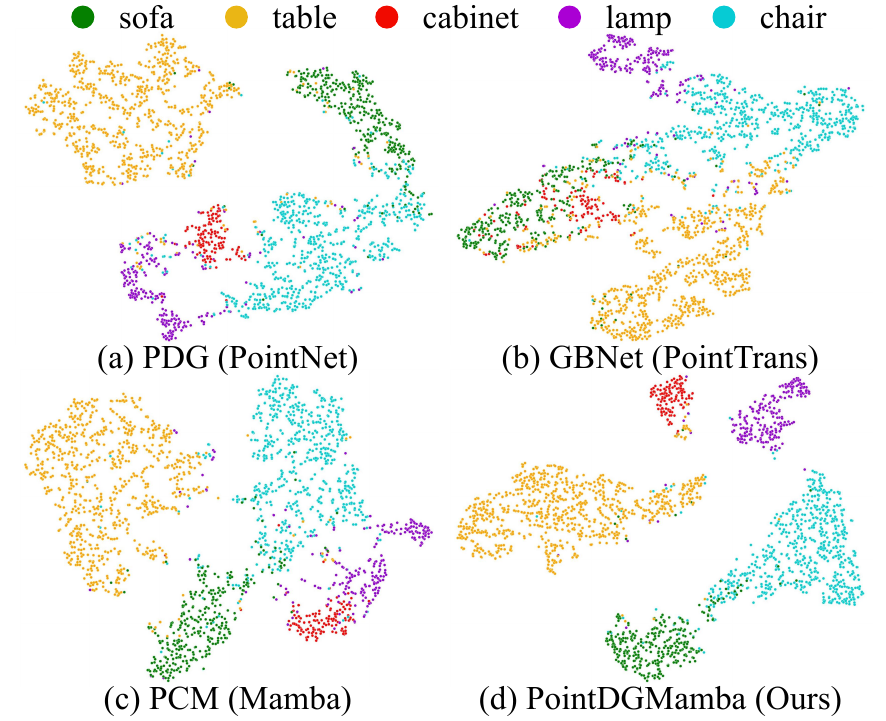}

\caption{Visualization of our PointDGMamba and other state-of-the-art methods using t-SNE, where they are tested on the ShapeNet-5(C) dataset of the PointDG-3to1 benchmark. Different colors represent different classes. 
}

\label{vis-compare}
\end{figure}

\begin{figure}[t]
\centering

\includegraphics[scale=0.492]{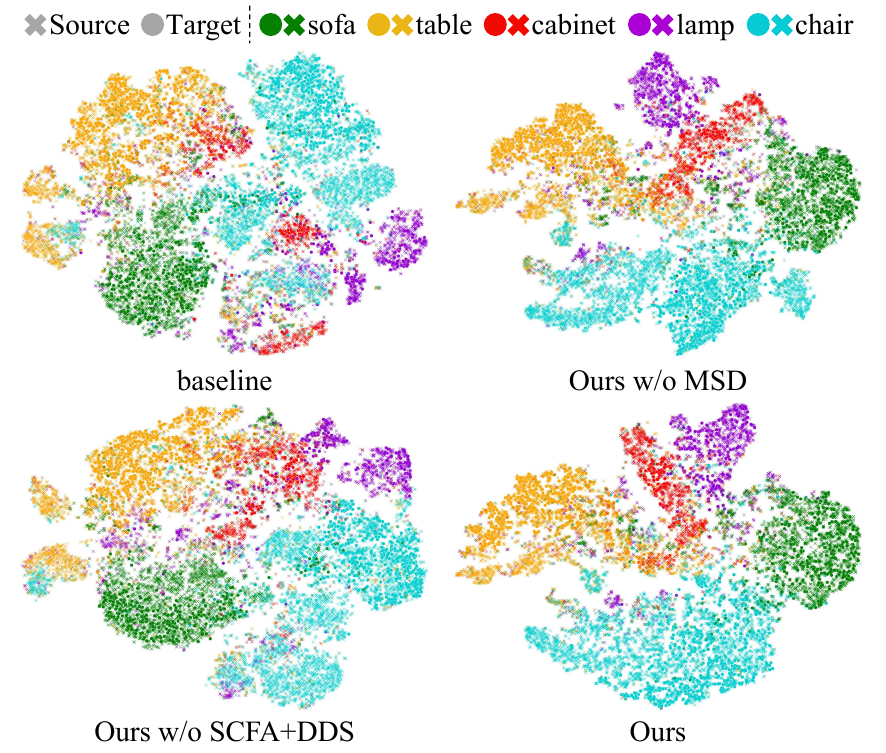}

\caption{Visualization on the distributions of ablations of our PointDGMamba. We also visualize the source and target domains, with marker \enquote{×} representing the entire source domain and circles representing the target domain. 
}
\label{vis-ablation}
\end{figure}

\subsection{Visualization and Analysis}

\noindent {\bf{Feature Visualization.}}
To more intuitively demonstrate the impact of PointDGMamba, we used t-SNE to visualize the features. We chose to visualize the testing set of ShapeNet-5(C) dataset on the PointDG-3to1 benchmark. This is because the testing sets of others have too few point clouds, or the number of point clouds of different classes varies by order of magnitude, which is not intuitive enough to verify our motivation. Figure~\ref{vis-compare} shows the feature visualization of the PointNet-based method PDG, Point Transformer-based method GBNet, Mamba-based method PCM, and our PointDGMamba. Specifically, PointDGMamba has achieved excellent characterization, manifested as stronger intra-class compactness and inter-class discrimination ability. Especially on the \enquote{cabinet (blue)} and \enquote{lamp (green)}, the difference between PointDGMamba is more obvious, with stronger compactness. These findings confirm the superiority of PointDGMamba in improving generalizability. 

In addition, we also visualized the impact of different modules on PointDGMamba. The SCFA and DDS modules have a causal relationship and must be used together. As shown in Figure~\ref{vis-ablation}, we visualize the point cloud features of the target and the source domains. 
From the perspective of the target domain, the model has poor inter-class discrimination ability when no modules are used. In the absence of MSD or SCFA+DDS, the inter-class discriminative ability of the model has been improved to some extent, but the compactness of some classes is poor, such as the \enquote{lamp (purple)}. When all modules are used together, our method can achieve optimal performance. From the source domain perspective, incomplete point clouds in ScanNet-5 make it difficult for the model to fully align point clouds of the same class in the feature space. After adding all modules, our method still allows point clouds of the same class in the source and target domains to stay closer in the feature space, such as \enquote{table (orange)} and \enquote{chair (Cyan)}.

\section{Conclusion}

We propose PointDGMamba, a novel State Space Model-based framework for domain generalizable point cloud classification. It excels in strong generalizability toward unseen domains and has the advantages of global receptive fields and efficient linear complexity. Specifically, Masked Sequence Denoising is presented to mitigate the adverse effects of noise accumulation during the serialization stage. Cross-domain Feature Aggregation and Dual-level Domain Scanning are designed to strengthen Mamba's effectiveness in learning domain-invariant features and avoiding unanticipated domain-specific information in the sequence data. In addition, we also proposed a benchmark PointDG-3to1 that includes more domains. Extensive experiments with analyses demonstrate the effectiveness and superiority of our PDGMmaba against state-of-the-art competitors.

\section{Acknowledgments}

This work was supported by the National Natural Science Foundation of China (No. 72192821), the National Natural Science Foundation of China (No. 62472282), the Fundamental Research Funds for the Central Universities (No. YG2023QNA35) and YuCaiKe [2023] Project Number (231111310300).

\bibliography{aaai25}


\twocolumn[
\begin{@twocolumnfalse}
	\section*{\centering{Supplementary Material of PointDGMamba: Domain Generalization of Point Cloud Classification via Generalized State Space Model}}
\end{@twocolumnfalse}
]

The supplementary materials provide further details about our proposed method and dataset. The structure of the supplementary materials is as follows:

\begin{itemize}
    \item \textbf{Section 1:} More Ablation Studies
    \begin{itemize}
        \item \textbf{Section 1.1:} Effects of Different Masking Strategies
        \item \textbf{Section 1.2:} Impacts of Different Feature Aggregation Mechanisms
        \item \textbf{Section 1.3:} Ablations on Different Scanning Strategies
    \end{itemize}
    
    \item \textbf{Section 2:} Computational Efficiency
    \item \textbf{Section 3:} Visualization of PointDG-3to1 Benchmark
    \item \textbf{Section 4:} Failure Cases
    \item \textbf{Section 5:} Details of Comparison Methods
    \item \textbf{Section 6:} Limitations and Future Work
\end{itemize}

\section{More Ablation Studies}
In our PointDGMamba, we design three key components, namely Masked Sequence Denoising (MSD), Sequence-wise Cross-domain Feature Aggregation (SCFA), and Dual-level Domain Scanning (DDS). To further explore their effectiveness, we provide more ablation studies to reveal the contribution of each module. For simplicity, all ablation experiments are conducted on the PointDA-10 benchmark.

\subsection{Effects of Different Masking Strategies}

Table \ref{supp-MSD} shows the effect of different masking schemes while keeping the SCFA and DSS unchanged.
Firstly, Randomly Masking means randomly selecting 5\% of sequence features for the masking, and the average performance is less desired, only achieving 69.75\%. 
Secondly, the Similarity Mask preserves the sequence features with similarity scores in the top 80\%, where the similarity score measures the similarity between sequence features and cross-domain features. Specifically, for each sequence in the sequence features, its similarity score is the sum of its similarity to each sequence in the cross-domain features. However, it cannot distinguish whether each sequence feature is noisy or not, and only achieved an accuracy of 70.98\%. Thirdly, IIM Mask~\cite{wang2022domain} is a masking method that enhances key information in features and suppresses less important features. It also cannot remove noise, with an accuracy of only 69.07\%. Finally, when using our MSD, the model was able to remove most of the noise and achieved the best generalization performance, with accuracy at least 3.87\% higher than the other three mask methods.

\begin{table}[t]
    \renewcommand\arraystretch{1.15} 
		\setlength{\tabcolsep}{1.25mm}
\begin{center}
    \begin{tabular}
		{ c|ccc| >{\columncolor{lightgray}}c}

		\hline
        \bf{Masking} & M,S*→S & M,S→S* & S,S*→M & \bf{Avg.} \\
        \hline
        Randomly Mask & 80.60 & 47.82 & 80.84 & 69.75 \\
        Similarity Mask & 81.40 & 50.64 & 80.90 & 70.98 \\
        IIM Mask & 81.14 & 47.17 & 78.91 & 69.07 \\
        \hline 
        Ours MSD & 84.33 & 52.83 & 87.38 & 74.85 \\
        \hline

    \end{tabular}
\end{center}

\caption{Effects of Different Masking Strategies.}
    \label{supp-MSD}
\end{table}

\begin{table}[t]
    \renewcommand\arraystretch{1.15} 
		\setlength{\tabcolsep}{0.4mm}
\begin{center}

		\begin{tabular}
				{ c|ccc| >{\columncolor{lightgray}}c}

		\hline
        \bf{Feature Aggregation} & M,S*→S & M,S→S* & S,S*→M & \bf{Avg.} \\
        \hline
        \small{Feature Summation} & 82.10 & 49.55 & 86.16 & 72.60 \\
        \small{Feature Concatenation} & 83.19 & 49.50 & 86.62 & 73.10 \\
        \small{Feature FDA} & 83.23 & 49.72 & 85.02 & 72.66 \\
        \hline 
        Ours SCFA & 84.33 & 52.83 & 87.38 & 74.85 \\
        \hline

		\end{tabular}
\end{center}

\caption{Impacts of Different Feature Aggregation Methods.}
    \label{supp-SCFA}
\end{table}

\begin{table}[t]
    \renewcommand\arraystretch{1.15} 
		\setlength{\tabcolsep}{1.3mm}
\begin{center}
		\begin{tabular}
				{ c|ccc| >{\columncolor{lightgray}}c}
    
		\hline
        \bf{Scan} & M,S*→S & M,S→S* & S,S*→M & \bf{Avg.} \\
        
        \hline
        Forward Scan & 83.39 & 49.91 & 85.32 & 72.87 \\
        Backward Scan & 83.09 & 50.25 & 86.52 & 73.29 \\
        Shuffle Scan & 83.59 & 49.94 & 84.23 & 72.59 \\
        
        \hline
        Ours DDS & 84.33 & 52.83 & 87.38 & 74.85 \\
        \hline

		\end{tabular}
\end{center}

\caption{Ablations on Different Scanning Strategies.}

    \label{supp-DDS}
\end{table}

\subsection{Impacts of Different Feature Aggregation Mechanisms}

Table~\ref{supp-SCFA} shows the impacts of different feature aggregation mechanisms.
As shown in the table, Feature Summation and Feature Concatenation directly introduce features without applying any domain generalization-related operations, resulting in their inability to improve generalization performance significantly. Feature FDA~\cite{wang2022domain} is a Fourier transform-based feature aggregation method that does not require the introduction of cross-domain features, resulting in the worst generalization performance. When using our SCFA, the model can learn more generalized features through cross-domain feature aggregation operations, achieving the best generalization performance.

\subsection{Ablations on Different Scanning Strategies}
Table~\ref{supp-DDS} studies the ablations on different scanning strategies. We used three different scanning methods: Forward Scan, Backward Scan, and Shuffle Scan. 
Forward Scan means scanning sequence features from the front to the back, while Backward Scan means scanning from the back to the front. 
Unlike them, Shuffle Scan first shuffles the feature sequence and then scans it sequentially. 
As shown in Table~\ref{supp-DDS}, these two rigid and fixed scanning methods of Forward Scan and Backward Scan inevitably introduce human bias and largely ignore domain-agnostic considerations, resulting in poor generalization performance. The scanning order of Shuffle Scan is too disordered and not conducive to generalization. 
In contrast, the model can achieve the best generalization performance when using DDS specifically designed to promote generalizability toward unseen domains.

\section{Computational Efficiency}

When using a state space model to process point cloud data, we need a network of a certain scale to store a large amount of state information, manifested in the number of Mamba stages and the dimension of sequence features. 
In Table~\ref{supp-scale}, we explore the impact of networks of different scales on generalization performance. Specifically, based on the original model of PointDGMamba, \emph{i.e.,} PointDGMamba (Tiny), we design PointDGMamba (Base) and PointDGMamba (Small), where PointDGMamba (Base) only contains two Mamba stages, and the feature dimension of PointDGMamba (Small) is reduced to 2/3 of its original size (192→128). As shown in Table \ref{supp-scale}, we report their generalization performance. Even PointDGMamba (Small) still outperforms the existing state-of-the-art methods Meta Set~\cite{huang2021metasets} by 1.79\%.

To evaluate the computational efficiency of our proposed PointDGMamba, we also compare it with the state-of-the-art methods such as GAST~\cite{zou2021geometry}, GBNet~\cite{qiu2021geometric}, SUG~\cite{huang2023sug}, and PCM~\cite{zhang2024point} from different perspectives, including model parameters (M), floating point operations per second (GFlops), inference time (ms), and generalization performance. As shown in Table~\ref{tab-cost}, our PointDGMamba achieves the highest generalization performance with relatively lower computational overhead.  PointDGMamba (Small) achieves good generalization performance despite a 51\% reduction in GLOPs.

\section{Visualization of PointDG-3to1 Benchmark}

Our proposed PointDG-3to1 benchmark includes four sub-datasets: ModelNet-5 (A), ScanNet-5 (B), ShapeNet-5 (C), and 3D-FUTURE-Completion (D). There are 5 shared classes in each dataset, including \enquote{cabinet}, \enquote{chair}, \enquote{lamp}, \enquote{sofa}, and \enquote{table}. Table \ref{benchmark-number} shows the number of point cloud testing sets and training sets for different classes in each sub-dataset, as well as the total number of point clouds.

\begin{figure*}[!ht]
\centering

\includegraphics[scale=0.58]{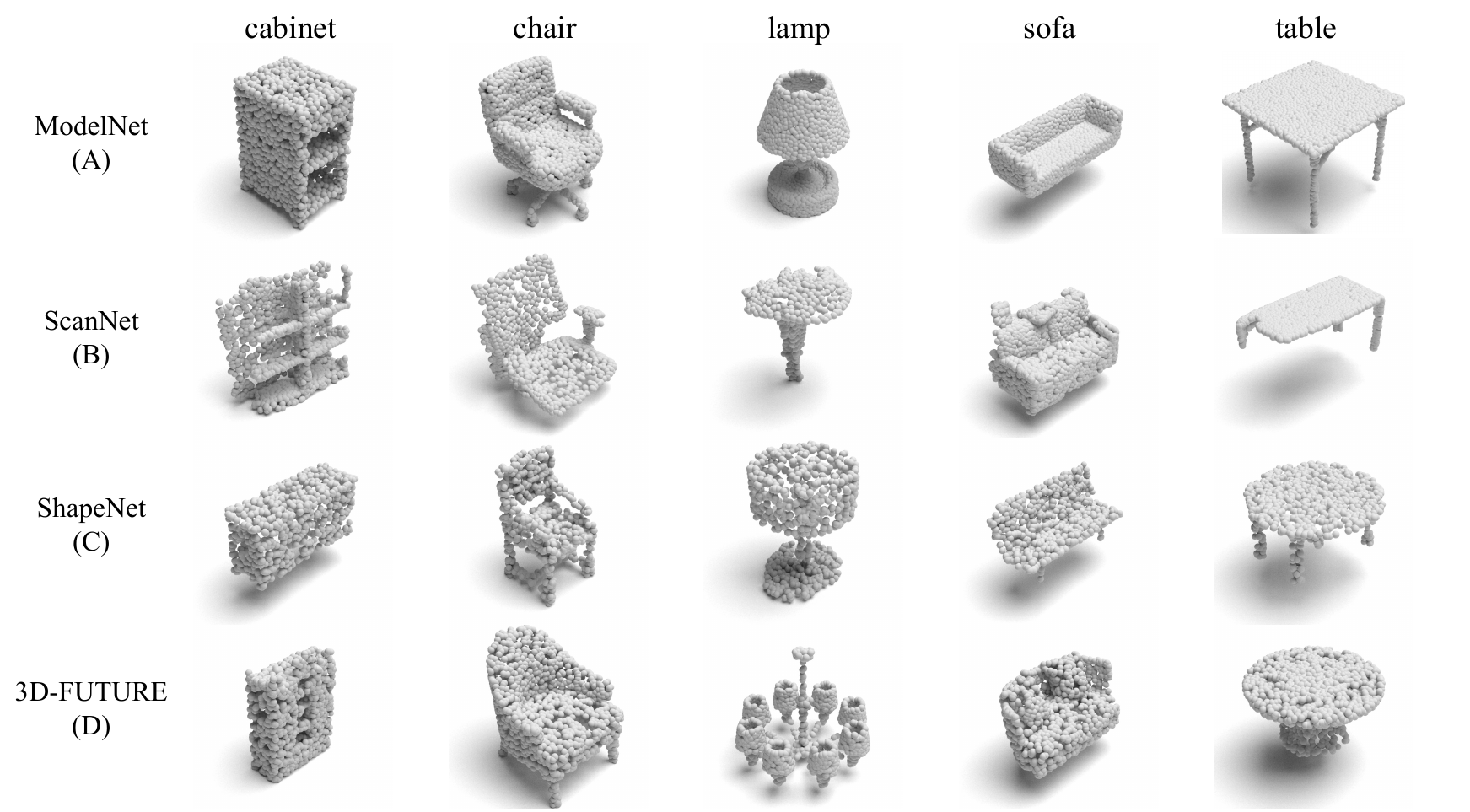}

\caption{Visualization of point clouds from different sub-datasets in our proposed PointDG-3to1 benchmark.}

\label{benchmark-vis}
\end{figure*}

We also visualize some point cloud samples in Figure \ref{benchmark-vis} to demonstrate the domain shifts between sub-datasets. The ModelNet dataset has the highest point cloud sample quality and visually compact point clouds among the four sub-datasets. The ScanNet dataset has the worst sample quality due to partially missing point clouds caused by object occlusion. The ShapeNet dataset still has some missing samples, such as \enquote{table}, but its quality is better than ScanNet. The point clouds in the 3D-FUTURE-Completion dataset appear not compact enough, and their quality visually falls between ModelNet and ShapeNet. Therefore, obvious domain shifts exist between these sub-datasets.

\section{Failure Cases}

Our proposed PointDGMamba model achieved the best generalization performance on both the PointDA-10 and PointDG-3to1 benchmarks. However, it still encounters some failure cases in classification, which vary depending on the sub-dataset. Taking the PointDG-3to1 benchmark as an example, the primary reason for failure in the ModelNet-5 (A), ShapeNet-5 (C), and 3D-FUTURE-Completion (D) datasets is that point clouds from different classes can be too similar in shape. For instance, some \enquote{cabinet} and \enquote{sofa} are both rectangular prisms, while single-person \enquote{sofa} and \enquote{chair} can appear very similar. In contrast, the main reason for failure cases in the ScanNet-5 (B) dataset is the presence of incomplete point clouds, such as a \enquote{table} with only one leg remaining or a \enquote{lamp} reduced to just a pole. These cases make it challenging for the model to correctly classify some point clouds.

\begin{table}[t]
    \renewcommand\arraystretch{1.15} 
		\setlength{\tabcolsep}{1.8mm}
\begin{center}

    \begin{tabular}
		{ c|ccc| >{\columncolor{lightgray}}c}

		\hline
        \bf{Scale} & M,S*→S & M,S→S* & S,S*→M & \bf{Avg.} \\
        \hline
        Ours-Base & 83.55 & 51.72 & 85.83 & 73.70 \\
        Ours-Small & 83.75 & 52.14 & 86.75 & 74.21 \\
        Ours-Tiny & 84.33 & 52.83 & 87.38 & 74.85 \\
        \hline
        
    \end{tabular}
\end{center}

\caption{The generalization performance of our PointDGMamba under different network scales.}

    \label{supp-scale}
\end{table}

\begin{table}[t]
    \renewcommand\arraystretch{1.1} 
		\setlength{\tabcolsep}{0.9mm}

		\begin{center}

		  \begin{tabular}
{c |ccc| >{\columncolor{lightgray}}c }
\hline
\bf{Method} & Params(M) & GFlops(G) & Time(ms) & \bf{Acc(\%)} \\
\hline

GAST & 75.36 & 2.17 & 23.13 & 69.61 \\
GBNet & 8.77 & 9.87 & 80.97 & 70.07 \\
SUG & 19.17 & 18.4 & 5.42 & 69.99 \\
PCM & 35.85 & 20.18 & 6.26 & 70.59 \\

\hline
Ours-Base & 7.72 & 3.76 & 2.10 & 73.68 \\
Ours-Small & 8.85 & 2.98 & 1.82 & 74.21 \\
Ours-Tiny & 13.09 & 6.08 & 3.35 & 74.85 \\
\hline

            \end{tabular}
        \end{center}

\caption{Comparison of computational efficiency between our method and existing methods. The testing was conducted on one NVIDIA 4090 GPU. }
\label{tab-cost}
\end{table}

\begin{table*}[t]
    \renewcommand\arraystretch{1.0} 
		\setlength{\tabcolsep}{1.3mm}
		\begin{center}
			\begin{tabular}{
                    >{\centering\arraybackslash}p{1.8cm} 
                    >{\centering\arraybackslash}p{1.1cm} 
                    >{\centering\arraybackslash}p{1.5cm} 
                    >{\centering\arraybackslash}p{1.5cm} 
                    >{\centering\arraybackslash}p{1.3cm} 
                    >{\centering\arraybackslash}p{1.2cm}
                    >{\centering\arraybackslash}p{1.3cm} 
                    >{\centering\arraybackslash}p{1.3cm} 
                    >{\centering\arraybackslash}p{1.5cm} }
                \midrule
				Dataset & Symbol & Partition & \textit{Cabinet} & \textit{Chair} & \textit{Lamp} & \textit{Sofa} & \textit{Table} & Total \\
				
                \midrule
				\multirow{2}{*}{ModelNet} & \multirow{2}{*}{A} & Train & 200 & 889 & 124 & 680 & 392 & 2285\\
                & & Test & 86 & 100 & 20 & 100 & 100 & 406\\
                \midrule
                \multirow{2}{*}{ScanNet} & \multirow{2}{*}{B} & Train & 650 & 2578 & 161 & 495 & 1037 & 4921 \\ 
                & & Test & 149 & 801 & 41 & 134 & 301 & 1426 \\ 
                \midrule
                \multirow{2}{*}{ShapeNet} & \multirow{2}{*}{C} & Train & 1076 & 4612 & 1620 & 2198 & 5876 & 15382 \\ 
                & & Test & 126 & 662 & 232 & 330 & 842 & 2192 \\
                \midrule
                \multirow{2}{*}{3D-FUTURE} & \multirow{2}{*}{D} & Train & 713 & 2034 & 1728 & 2193 & 2052 & 8720 \\ 
                & & Test & 80 & 226 & 193 & 244 & 228 & 971 \\   
				\midrule
			\end{tabular}
	\end{center}

\caption{The numbers of point clouds of different classes in each dataset in our proposed PointDG-3to1 benchmark.}

\label{benchmark-number}
\end{table*}

\section{Details of Comparison Methods}
As some of the comparison methods we have chosen are not originally designed for DG, we clarify their details that adapt to the DG setting. For the methods specifically designed for DG, we followed the original settings during the experiment and did not make any modifications, such as PDG~\cite{wei2022learning} and MetaSets~\cite{huang2021metasets}. 
For methods not designed for DG, such as PCM~\cite{zhang2024point} and GBNet~\cite{qiu2021geometric}, we have made certain modifications to ensure they 
strictly follow the protocol of PDG and MetaSets. Specifically, we train these models on the training set of the entire source domain and test them on the test set of the target domain. The classification accuracy on the target domain test set will serve as the final indicator for evaluating generalization performance.
For example, in the case of ABC→D, we combine the training set of ABC into a single overall training set and train the model on it, then test it on the testing set of D. For methods specifically designed for DA, we also follow their original settings. In the data preprocessing stage of all methods, we use the same normalization and jitter operations to process the training data, while only normalization operations are used for the test data.

\section{Effect of Module Insertion Position}
Table ~\ref{tab4} illustrates the impact of inserting the presented module into the $i$-th position of PointDGMamba \emph{i.e.,} between the $(i-1)$-th and $i$-th Mamba stages.
From the tale, we have the following observations: 1)
When all modules are inserted into position 1, the model has the best generalization performance. 2) When inserting the proposed module at positions 2 and 3, the generalization effect of the model gradually decreases. This is because when being closer to the input data, the low-level features extracted by the model are more generalizable. The closer the position to the classification head becomes, the more discriminated features the model extracts. 3) In addition, we observed that the generalization performance further decreases when the modules are separated and inserted into different positions. This indicates that sequence-wise cross-domain feature aggregation is more advantageous for generalization immediately after denoising.

\begin{table}[t]
    \renewcommand\arraystretch{1.02} 
    \setlength{\tabcolsep}{0.9mm}
\begin{center}
        \begin{tabular}{c|c|ccc| >{\columncolor{lightgray}}c}
\hline
\multicolumn{2}{c|}{\bf{Position}} & \multirow{2}{*}{M,S*→S} & \multirow{2}{*}{M,S→S*} & \multirow{2}{*}{S,S*→M} &  \\
\cline{1-2}
\bf{MSD} & \bf{SCFA+DDS} &  &  &  & \multirow{-2}{*}{\bf{Avg.}}\\
\hline
3 & 3 & 83.91 & 51.67 & 86.53 & 74.04  \\
2 & 3 & 82.98 & 51.50 & 85.52 & 73.33  \\
2 & 2 & 83.81 & 52.31 & 86.81 & 74.31  \\
1 & 3 & 83.56 & 52.25 & 85.75 & 73.85  \\
1 & 2 & 83.82 & 52.06 & 86.11 & 74.00  \\
\hline
1 & 1 & 84.33 & 52.83 & 87.38 & 74.85  \\
\hline
\end{tabular}
\end{center}

\caption{Effect of Module Insertion Position.}

\label{tab4}
\end{table}

\section{Limitations and Future Work}
PointDGMamba successfully introduced Mamba into DG PCC and outperformed existing CNN-based and ViT-based methods, achieving the best domain generalization performance. 
However, there are still some shortcomings that need to be further explored. 
Due to Mamba's scan-based computation method, it is necessary to crop longer point cloud sequences during the training process to reduce training time effectively. However, this clipping method may affect the representativeness of the features learned by the model. How to extract key features that reflect the entire point cloud from the clipped point cloud is still a problem that needs further research. In addition, there are many directions worth exploring when introducing PointDGMamba into point cloud segmentation tasks, especially considering that the scale of point cloud data in segmentation scenarios is usually very large. How to effectively process and utilize these large-scale point cloud data will be an important challenge in the future. In the next step of our work, we will investigate how to address this challenge.



\end{document}